\def\year{2021}\relax
\documentclass[letterpaper]{article} 
\usepackage{aaai21}  
\usepackage{times}  
\usepackage{helvet} 
\usepackage{courier}  
\usepackage[hyphens]{url}  
\usepackage{graphicx} 
\usepackage{natbib}  
\usepackage{caption} 
\usepackage{subcaption}
\usepackage{amsmath, amssymb}
\usepackage{algorithm} 
\usepackage{algpseudocode} 
\usepackage{todonotes}

\title{Value-Decomposition Multi-Agent Actor-Critics}
\author{

    Jianyu Su, Stephen Adams, Peter A. Beling
    \\
}
\affiliations{

    University of Virginia\\

    151 Engineer's Way \\
    Charlottesville, Virginia, 22904\\
    \{js9wv, sca2c, pb3a\}@virginia.edu

}

\begin{document}

\maketitle

\begin{abstract}
The exploitation of extra state information has been an active research area in multi-agent reinforcement learning (MARL). QMIX represents the joint action-value using a non-negative function approximator and achieves the best performance on the StarCraft II micromanagement testbed, a common MARL benchmark. However, our experiments demonstrate that, in some cases, QMIX performs sub-optimally with the A2C framework, a training paradigm that promotes algorithm training efficiency. To obtain a reasonable trade-off between training efficiency and algorithm performance, we extend value-decomposition to actor-critic methods that are compatible with A2C and propose a novel actor-critic framework, value-decomposition actor-critic (VDAC). We evaluate VDAC on the StarCraft II micromanagement task and demonstrate that the proposed framework improves median performance over other actor-critic methods. Furthermore, we use a set of ablation experiments to identify the key factors that contribute to the performance of VDAC.
\end{abstract}

\noindent Many complex sequential decision making problems that involve multiple agents can be modeled as multi-agent reinforcement learning (MARL) problems, e.g. the coordination of semi-autonomous or fully autonomous vehicles \cite{Hu2019InteractionawareDM} and the coordination of machines in a product line \cite{choo2017health}. A fully centralized controller that applies single-agent reinforcement learning will suffer from the exponential growth of the action space with the number of agents in the system.  Learning decentralized policies that condition on the local observation history of individual agents is a viable way to attenuate this problem.  Furthermore, partial observability and communication constraints, two common obstacles in multi-agent settings, also necessitate the use of decentralized policies. 

In a laboratory or simulated setting, decentralized policies can be learned in a centralized fashion via enabling communication among agents or granting access to additional global state information. This \textit{centralized training and decentralized execution} (CTDE) paradigm has attracted the attention of researchers. However, it remains an open research question how to best exploit centralized training.  In particular, it is not obvious how to utilize joint action-value or global state value to train decentralized policies.

Breakthroughs in Q-learning have been made using joint action-value factorization techniques. \textit{Value-decomposition networks} (VDN) represent joint action-value as a summation of local action-value conditioned on individual agents' local observation history \cite{sunehag2017value}. In \cite{rashid2018qmix}, a more general case of VDN is proposed using a mixing network that approximates a broader class of monotonic functions to represent joint action-values called QMIX. In \cite{son2019qtran}, a more complex factorization framework three modules, called QTRAN, is introduced and shown to have good performance on a range of cooperative tasks. While QMIX reports the best performance on the StarCraft micromanagement testbed \cite{samvelyan2019starcraft}, we find that QMIX, in some StarCraft II compositions, has issues learning good policies that can consistently defeat enemies when using the A2C training paradigm \cite{mnih2016asynchronous}, which was originally introduced to enable algorithms to be executed efficiently.

On the other hand, on-policy actor-critic methods, such as \textit{counterfactual multi-agent} (COMA) \cite{foerster2018counterfactual}, can leverage the A2C framework to improve training efficiency at the cost of performance. \cite{samvelyan2019starcraft} point out that there is a performance gap between the state-of-the-art actor-critic method, COMA, and QMIX on the StarCraft II micromanagement testbed.
 
To bridge the gap between multi-agent Q-learning and multi-agent actor-critic methods, as well as offer a reasonable trade-off between training efficiency and algorithm performance, we propose a novel actor-critic framework called value-decomposition actor-critic (VDAC). Let $V^a,\forall a\in \{1,\dots,n\}$ denote the local state value that is conditioned on agent $a$'s local observation, and let $V_{tot}$ denote the global state-value that is conditioned on the true state of the environment. VDAC takes an actor-critic approach but adds local critics, which share the same network with the actors and estimate the local state values $V^a$.  The central critic learns the global state value $V_{tot}$. The policy is trained by following a gradient dependent on the central critic. Further, we examines two approaches for calculating $V_{tot}$.

VDAC is based on three main ideas.  First, unlike QMIX, VDAC is compatible with a A2C training framework that enables game experience to be sampled efficiently. This is due to the fact that multiple games are rolled out independently during training. Second, similar to QMIX, VDAC enforces the following relationship between local state-values $V^a$ and the global state-value $V_{tot}$:
\begin{equation} \label{eq: mono1}
    \frac{\partial V_{tot}}{\partial V^a} \geq 0, \quad \forall a\in \{1,\dots,n\}.
\end{equation}
This idea is related to \textit{difference rewards} \cite{wolpert2002optimal}, in which each agent learns from a shaped reward that compares the global reward to the reward received when that agent’s action is replaced with a default action. \textit{Difference rewards} require that any action that improves an agent's local reward also improves the global reward, which implies the monotonic relationship between shaped local rewards and the global reward. While COMA (also inspired by \textit{difference rewards}) focuses on customizing shaped rewards $r^a$ from the global reward $r_{tot}$ in a pairwise fashion $r^a=f(r_{tot})$, VDAC represents the global reward by all agents' shaped rewards $r_{tot}=f(r^1,...,r^n)$ . Third, VDAC is trained by following a rather simple policy gradient that is calculated from a temporal-difference (TD) advantage. We theoretically demonstrate that the proposed method is able to converge to a local optimum by following this policy gradient. Despite the fact that TD advantage policy gradients and COMA gradients are both unbiased estimates of a vanilla multi-agent policy gradients, our empirical study favors TD advantage policy gradients over COMA policy gradients.

This study strives to answer the following research questions:
\begin{quote}
\begin{itemize}
    \item \textbf{Research question 1}: Is the TD advantage gradient sufficient to optimize multi-agent actor-critics when compared to a COMA gradient?
    \item \textbf{Research question 2}: Does applying state-value factorization improve the performance of actor-critics?
    \item \textbf{Research question 3}: Does VDAC provide a reasonable trade-off between training efficiency and algorithm performance when compared to QMIX?
    \item \textbf{Research question 4}: What are the factors that contribute to the performance of the proposed VDAC?
\end{itemize}
\end{quote}


\section{Related Work}\label{sec:related}
MARL has benefited from recent developments in deep reinforcement learning, with the field moving away from tabular methods \cite{bu2008comprehensive} to deep neural networks \cite{foerster2018counterfactual}. Our work is related to recent advances in CTDE deep multi-agent reinforcement learning. 

The degree of training centralization varies in the literature on MARL. \textit{Independent Q-learning} (IQL) \cite{tan1993multi} and its deep neural network counterpart \cite{tampuu2017multiagent} train an independent Q-learning model for each agent. Those that attempt to directly learn decentralized policies often suffer from the non-stationarity of the environment induced by agents simultaneously learning and exploring. \cite{foerster2017stabilising,usunier2016episodic} attempt to stabilize learning under the decentralized training paradigm. \cite{gupta2017cooperative} propose a training paradigm that alternates between centralized training with global rewards and decentralized training with shaped rewards.

Centralized methods, by contrast, naturally avoid the non-stationary problem at the cost of scalability. COMA \cite{foerster2016learning}, takes advantage of CTDE, where actors are updated by following policy gradients that are tailored by their contributions to the system. \textit{Multi-agent deep deterministic policy gradient} (MADDPG) \cite{lowe2017multi} extends \textit{deep deterministic policy gradient} (DDPG) \cite{lillicrap2015continuous} to mitigate the issue of high variance gradient estimates exacerbated in multi-agent settings. Based on MADDPG, \cite{wei2018multiagent} propose multi-agent soft Q-learning in continuous action spaces to tackle the issue of \textit{relative overgeneralization}. \textit{Probabilistic recursive reasoning} \cite{wen2019probabilistic} is a method that uses a probabilistic recursive reasoning policy gradient that enables agents to recursively reason what others believe about their own beliefs.

More recently, value-based methods, which lie between the extremes of IQL and COMA, have shown great success in solving complex multi-agent problems. VDN \cite{sunehag2017value}, which represents joint-action value function as a summation of local action-value function, allows for centralized learning. However, it does not make use of extra state information. QMIX \cite{rashid2018qmix} utilizes a non-negative mixing network to represent a broader class of value-decomposition functions. Furthermore, additional state information is captured by hypernetworks that output parameters for the mixing network. QTRAN \cite{son2019qtran} is a generalized factorization method that can be applied to environments that are free from structural constraints. Other works, such as CommNet \cite{foerster2016learning}, TarMAC \cite{das2019tarmac}, ATOC \cite{jiang2018learning}, MAAC \cite{iqbal2019actor}, CCOMA \cite{su2020counterfactual} and BiCNet\cite{peng2017multiagent} exploit inter-agent communication.

The proposed VDAC method is similar to QMIX and VDN in that it utilizes value-decomposition. However, VDAC is a policy-based method that decomposes global state-values whereas QMIX and VDN, which decompose global action-values, belong to the Q-learning family. \cite{nguyen2018credit} address credit-assignment issue, however, under a different MARL setting, $\mathbb{C}$Dec-POMDP. COMA, which is also a policy gradient method inspired by \textit{difference rewards} and has been tested on StarCraft II micromanage games, represents the work most closely related to this paper.

\section{Background}\label{sec:technical}
\textbf{Decentralized Partially Observable Markov Decision Processes (Dec-POMDPs)}: 
Consider a fully cooperative multi-agent task with $n$ agents. Each agent identified by $a \in A \equiv \{1,\dots,n\}$ takes an action $u^a \in U$ simultaneously at every timestep, forming a joint action $\mathbf{u} \in \mathbf{U} \equiv U^a, \forall a \in \{1,\dots,n\}$. The environment has a true state $s \in S$, a transition probability function $P(s'|s,\mathbf{u}):S\times \mathbf{U} \times S\rightarrow S$, and a global reward function $r(s,\mathbf{u}):S\times \mathbf{U} \rightarrow \mathbb{R}$. In the partial observation setting,  each agent draws an observations $z \in Z$ from the observation function $O(S, A): S \times A \rightarrow Z$. Each agent conditions a stochastic policy $\pi(u^a|\tau^a):T \times U \rightarrow [0, 1]$ on its observation-action history $\tau^a \in T\equiv Z \times U$. Throughout this paper, quantities in bold represent joint quantities over agents, and bold quantities with the superscript $-a$ denote joint quantities over agents other than a given agent $a$. MARL agents aim to maximize the discounted return $R_t=\sum_{l=1}^{\infty}\gamma ^l r_{t+l}$. The joint value function $V^{\mathbf{\pi}}(s_t)=\mathbb{E}[R_t|s_t=s]$ is the expected return for following the joint policy $\mathbf{\pi}$ from state $s$. The value-action function $Q^{\pi}(s,\mathbf{u})  =\mathbb{E}[R_t|s_t=s,\mathbf{u}]$ defines the expected return for selecting joint action $\mathbf{u}$ in state $s$ and following the joint policy $\mathbf{\pi}$.

\textbf{Single-Agent Policy Gradient Algorithms}: Policy gradient methods adjust the parameters $\theta$ of the policy in order to maximize the objective $J(\theta)=\mathbb{E}_{s\sim p^{\pi},u\sim \pi}[R(s,u)]$ by taking steps in the direction of $\nabla J(\theta)$. The gradient with respect to the policy parameters is $\nabla_{\theta} J(\theta)=\mathbb{E}_{\pi}[\nabla_{\theta} \log\pi_{\theta}(a|s)Q_{\pi}(s,u)]$, where $p^{\pi}$ is the state transition by following policy $\pi$, and $Q_{\pi}(s,u)$ is an action-value. 


To reduce variations in gradient estimates, a baseline $b$ is introduced. In actor-critic approaches \cite{konda2000actor}, an actor is trained by following gradients that are dependent on the critic. This yields the advantage function $A(s_t,u_t)=Q(s_t,u_t)-b(s_t)$, where $b(s_t)$ is the baseline ($V(s_t)$ or another constant is commonly used as the baseline). TD error $r_t+\gamma V(s_{t+1})-V(s_t)$, which is an unbiased estimate of $Q(s_t,u_t)$, is a common choice for advantage functions. In practice, a TD error that utilizes an n-step return $\sum_{i=0}^{k-1}\gamma^i r_t+\gamma^k V(s_{t+k})-V(s_t)$ yields good performance \cite{mnih2016asynchronous}.

\textbf{Multi-Agent Policy Gradient (MAPG) Algorithms}: Multi-agent policy gradient methods are extensions of policy gradient algorithms with a policy $\pi_{\theta_a}(u^a|o^a), a \in \{1,\cdots,n\}$ . Compared with policy gradient methods, MAPG faces the issues of high variance gradient estimates \cite{lowe2017multi} and credit assignment \cite{foerster2018counterfactual}. Perhaps the simplest multi-agent gradient can be written as:
\begin{equation} \label{eq: 2}
    \nabla_{\theta} J=\mathbb{E}_{\mathbf{\pi}} \left[\sum_a \nabla_{\theta} \log\mathbf{\pi}_{\theta}(u^a|o^a)Q_{\mathbf{\pi}}(s,\mathbf{u}) \right].
\end{equation}

Multi-agent policy gradients in the current literature often take advantage of CTDE by using a central critic to obtain extra state information $s$, and avoid using the vanilla multi-agent policy gradients (Equation \ref{eq: 2}) due to high variance. For instance, \cite{lowe2017multi} utilize a central critic to estimate $Q(s,(a_1,\dots,a_n))$ and optimize parameters in actors by following a multi-agent DDPG gradient, which is derived from Equation \ref{eq: 2}:
\begin{equation} \label{eq: 3}
    \nabla_{\theta_a} J=\mathbb{E}_{\mathbf{\pi}}[ \nabla_{\theta_a} \mathbf{\pi}(u^a|o^a)\nabla_{u^a}Q_{{u}^a}(s,\mathbf{u})|_{u^a=\mathbf{\pi}(o^a)}].
\end{equation}
Unlike most actor-critic frameworks, \cite{foerster2018counterfactual} claim to solve the credit assignment issue by applying the following counterfactual policy gradients:
\begin{equation} \label{eq: 4}
    \nabla_{\theta} J=\mathbb{E}_{\mathbf{\pi}} \Bigg[\sum_a \nabla_{\theta}  \log \pi(u^a|\tau^a) A^a(s,\mathbf{u}) \Bigg],
\end{equation}
\noindent where $A^a(s,\mathbf{u})=Q_{\mathbf{\pi}}(s,\mathbf{u})-\sum_{u^a}\pi_{\theta}(u^a|\tau^a)Q_{\mathbf{\pi}}^a(s,(\mathbf{u}^{-a},u^a))$ is the counterfactual advantage for agent $a$. Note that \cite{foerster2018counterfactual} argue that the COMA gradients provide agents with tailored gradients, thus achieving credit assignment. At the same time, they also prove that COMA is a variance reduction technique. 

\section{Methods}\label{sec:method}

In addition to the previously outlined research questions, our goal in this work is to derive RL algorithms under the following constraints: (1) the learned policies are conditioned on agents' local action-observation histories (the environment is modeled as Dec-POMDP), (2) a model of the environment dynamics is unknown (i.e. the proposed framework is task-free and model-free), (3) communication is not allowed between agents (i.e. we do not assume a differentiable communication channel such as \cite{das2019tarmac}), and (4) the framework should enable parameter sharing among agents (namely, we do not train different models for each agent as is done in \cite{tan1993multi}). A method that met the above criteria would constitute a general-purpose multi-agent learning algorithm that could be applied to a range of cooperative environments, with or without communication between agents. Hence, the following methods are proposed.

\begin{table*}[t] 
\centering
\caption{Actor-Critics studied.}
\begin{tabular}{l l l c l c c}
\hline
 Algorithm&  & &Central Critic & & Value Decomposition & Policy Gradients\\
\hline
IAC \cite{foerster2018counterfactual} & & &No & &- &TD advantage\\
VDAC-sum & & &Yes & &Linear& TD advantage\\
VDAC-mix & & &Yes & &Non-linear& TD advantage\\
Naive Critic & & &Yes & &-&TD advantage\\
COMA \cite{foerster2018counterfactual} & & &Yes & &- &COMA advantage\\
\hline
\hline
\label{table: tj}
\end{tabular}
\end{table*}

\subsection{Naive Central Critic Method}
A naive central critic (naive critic) is proposed to answer the first research question: is a simple policy gradient sufficient to optimize multi-agent actor-critic methods. Naive critic's central critic shares a similar structure with COMA's critic. It takes $(s_t,u_{t-1})$ as the input and outputs $V(s)$. Actors follow a rather simple policy gradient, a TD advantage policy gradient that is common in the RL literature given by:
\begin{equation} \label{eq: pg}
\nabla_{\theta} J= \mathbb{E}_{\mathbf{\pi}} \Bigg[\sum_a \nabla_{\theta}  \log \pi(u^a|\tau^a) \big(Q(s, \mathbf{u})-V(s)\big)  \Bigg],
\end{equation}

\noindent where $Q(s, \mathbf{u})=r+\gamma V(s')$. In the next section, we will demonstrate that policy gradients taking the form of Equation \ref{eq: pg} under our proposed actor-critic frameworks are also  unbiased estimates of the naive multi-agent policy gradients. The pseudo code is listed in Appendix.



\subsection{Value Decomposition Actor-Critic}
\textit{Difference rewards} enable agents to learn from a shaped reward $D^a=r(s,\mathbf{u})-r(s,(\mathbf{u}^{-a},c^a))$ that is defined by a reward change incurred by replacing the original action $u^a$ with a default action $c^a$. Any action taken by agent $a$ that improves $D^a$ also improves the global reward $r(s,\mathbf{u})$ since the second term in the difference reward equation does not depend on $u^a$.  Therefore, the global reward $r(s,\mathbf{u})$ is monotonically increasing with $D^a$. Inspired by \textit{difference rewards}, we propose to decompose state value $V_{tot}(s)$ into local states $V^a(o^a)$ such that the following relationship holds:
\begin{equation} \label{eq: mono}
    \frac{\partial V_{tot}}{\partial V^a} \geq 0, \quad \forall a\in \{1,\dots,n\}.
\end{equation}
With Equation \ref{eq: mono} enforced, given that the other agents stay at the same local states by taking $\mathbf{u}^{-a}$, any action $u^a$ that leads agent $a$ to a local state $o^a$ with a higher value will also improve the global state value $V_{tot}$. 

Two variants of value-decomposition that satisfy Equation \ref{eq: mono}, VDAC-sum and VDAC-mix, are studied. 

\subsubsection{VDAC-sum}

\begin{figure}[t]
\centering
\includegraphics[scale=0.3]{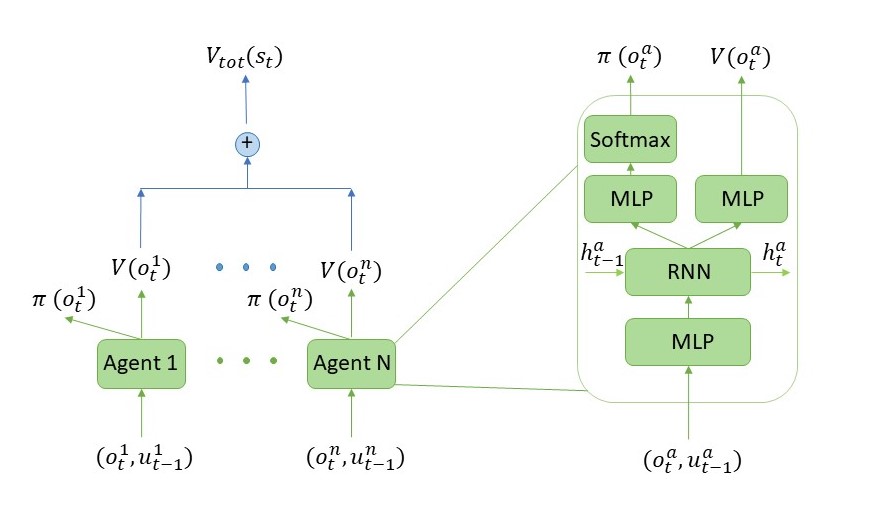} 
\caption{VDAC-sum}
\label{fig: vdn}
\end{figure}

In VDAC-sum, the total state value $V_{tot}(s)$ is a summation of local state values $V^a(o^a)$: $V_{tot}(s) = \sum_a V^a(o^a)$.  This linear representation is sufficient to satisfy Equation \ref{eq: mono}. VDAC-sum's structure is shown in Figure \ref{fig: vdn}. Note that the actor outputs both $\pi_{\theta}(o^a)$ and $V_{\theta_v}(o^a)$. This is done by sharing non-output layers between distributed critics and actors. In this paper, $\theta_v$ denotes the distributed critics' parameters and $\theta$ denotes the actors' parameters for generality. The distributed critic is optimized by minibatch gradient descent to minimize the following loss:
\begin{equation}
\begin{split}
    L_{t}(\theta_v) &=\bigg(y_{t}-V_{tot}(s_t)\bigg)^2\\
                  &=\bigg(y_{t}-\sum_aV_{\theta_v}(o^a_t)\bigg)^2,\\
\end{split}
\end{equation}

\noindent where $y_{t}=\sum_{i=t}^{k-t-1}\gamma^{i}r_i+\gamma^{(k-t)}V_{tot}(s_{k})$ is bootstrapped from the last state $s_{k}$, and $k$ is upper-bounded by $T$. 

The policy network is trained using the following policy gradient $g=\mathbb{E}_{\mathbf{\pi}} [\sum_a \nabla_{\theta}  \log \pi(u^a|\tau^a) A(s,\mathbf{u}) ]$, where $A(s,\mathbf{u})=r+\gamma V(s') -V(s)$ is a simple TD advantage. Similar to independent actor-critic (IAC), VDAC-sum does not make full use of CTDE in that it does not incorporate state information during training. Furthermore, it can only represent a limited class of centralized state-value functions. 

\subsubsection{VDAC-mix} 
\begin{figure}[t]
\centering
\includegraphics[scale=0.25]{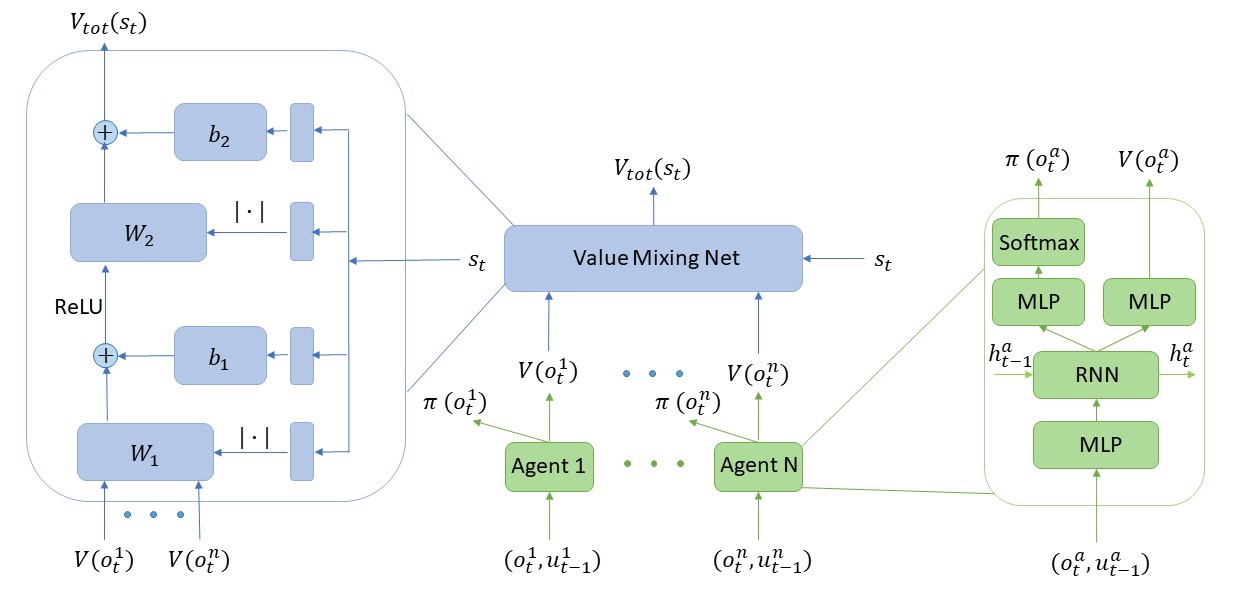} 
\caption{VDAC-vmix}
\label{fig: vmix}
\end{figure}
To generalize the representation to a larger class of monotonic functions, we utilize a feed-forward neural network that takes input as local state values $V_{\theta}(o^a), \forall a \in \{1,\dots,n\}$ and outputs the global state value $V_{tot}$. To enforce Equation \ref{eq: mono}, the weights (not including bias) of the network are restricted to be non-negative. This allows the network to approximate any monotonic function arbitrarily well \cite{dugas2009incorporating}.

The weights of the mixing network are produced by separate hypernetworks \cite{ha2016hypernetworks}. Following the practice in QMIX \cite{rashid2018qmix}, each hypernetwork takes the state $s$ as an input and generates the weights of one layer of the mixing network. 
Each hypernetwork consists of a single linear layer. An absolute activation function is utilized in the hypernetwork to ensure that the outputted weights are non-negative. The biases are not restricted to being non-negative. Hence, the hypernetworks that produce the biases do not apply an absolute non-negative function. The final bias is produced by a 2-layer hypernetwork with a ReLU activation function following the first layer. Finally, the hypernetwork outputs are reshaped into a matrix of appropriate size. Figure \ref{fig: vmix} illustrates the mixing network and the hypernetworks.

The whole mixing network structure (including hypernetworks) can be seen as a central critic. Unlike critics in \cite{foerster2018counterfactual}, this critic takes local state values $V^a(o^a), \forall a \in \{1,\dots,n\}$ as additional inputs besides global state $s$. Similar to VDAC-sum, the distributed critics are optimized by minimizing the following loss:
\begin{equation}
\begin{split}
    L_{t}(\theta_v) &=\bigg(y_{t}-V_{tot}(s_t)\bigg)^2\\
                  &=\bigg(y_{t}-f_{mix}(V_{\theta_v}(o^1_t),\dots,V_{\theta_v}(o^n_t))\bigg)^2,\\
\end{split}
\end{equation} 

\noindent where $f_{mix}$ denotes the mixing network. Let $\theta^c$ denote parameters in the hypernetworks. The central critic is optimized by minimizing the same loss $ L_{t}(\theta^c) =(y_{t}-V_{tot}(s_t))$.  The policy network is updated by following the same policy gradient in Equation \ref{eq: pg}. The pseudo code is provided in Appendix.

\subsubsection{Convergence of VDAC frameworks}
 \cite{foerster2018counterfactual} establish the convergence of COMA based on the convergence proof of single-agent actor-critic algorithms \cite{konda2000actor,sutton2000policy}. In the same manner, we utilize the following lemma to substantiate the convergence of VDACs to a locally optimal policy.

\textbf{Lemma 1}: For a VDAC algorithm with a compatible TD(1) critic following a policy gradient
$$g_k=\mathbb{E}_{\mathbf{\pi}} \Bigg[\sum_a \nabla_{\theta_k} \log \pi(u^a|\tau^a) A(s,\mathbf{u})) \Bigg],$$
at each iteraction $k$,
$\lim \text{inf}_k ||\nabla J||=0 \quad w.p. 1.$

\textit{Proof} The VDAC gradient is given by:
\begin{equation}
    g=\mathbb{E}_{\mathbf{\pi}} \Bigg[\sum_a \nabla_{\theta}  \log \pi(u^a|\tau^a) A(s,\mathbf{u}) \Bigg],
\end{equation}

\noindent $A(s,\mathbf{u}) = Q(s,\mathbf{u})-V_{tot}(s)$. We first consider the expected distribution of the baseline $V_{tot}$:
\begin{equation}
\begin{split}
    g_b &=-\mathbb{E}_{\mathbf{\pi}} \Bigg[\sum_a \nabla_{\theta} \log \pi(u^a|\tau^a) V_{tot}(s) \Bigg] \\ 
     &=-\mathbb{E}_{\mathbf{\pi}} \Bigg[\nabla_{\theta}  \log \prod_{a}\pi(u^a|\tau^a) V_{tot}(s) \Bigg],
\end{split}
\end{equation}
\noindent where the distribution $\mathbb{E}_{\pi}$ is with respect to the state-action distribution induced by the joint policy $\mathbb{\pi}$. 
Writing the joint policy as a product of independent actors $\mathbf{\pi}(\mathbf{u}|s)=\prod_a \pi(u^a|\tau^a)$.  The total value does not depend on agent actions and is given by $V_{tot}(s)=f(V_1(o^1),\dots,V_n(o^n))$ where $f$ is a non-negative function. This yields a single-agent actor-critic baseline: $g_b=-\mathbb{E}_{\mathbf{\pi}} [\nabla_{\theta} \log \mathbf{\pi}(\mathbf{u}|s) V_{tot}(s) ]$.

Now let $d^{\pi(s)}$ be the discounted ergodic state distribution as defined by \cite{sutton2000policy}:
\begin{equation} \label{eq2}
\begin{split}
 g_b & = -\sum_{s}d^{\mathbf{\pi}(s)} \sum_{\mathbf{u}} \nabla_{\theta}\log \mathbf{\pi}(\mathbf{u}|s) V_{tot}(s)  \\ 
 & = -\sum_{s}d^{\mathbf{\pi}(s)}V_{tot}(s) \nabla_{\theta}\sum_{\mathbf{u}} \log \mathbf{\pi}(\mathbf{u}|s)\\
 & = -\sum_{s}d^{\mathbf{\pi}(s)}V_{tot}(s) \nabla_{\theta} 1 \\
 & =0  \\
\end{split}
\end{equation}

The remainder of the gradient is given by:
\begin{equation}
\begin{split}
    g &=\mathbb{E}_{\mathbf{\pi}} \Bigg[\sum_a \nabla_{\theta} \log \pi(u^a|\tau^a) Q(s, \mathbf{u}) \Bigg] \\ 
     &=\mathbb{E}_{\mathbf{\pi}} \Bigg[\nabla_{\theta}  \log \prod_{a}\pi(u^a|\tau^a) Q(s, \mathbf{u}) \Bigg],
\end{split}
\end{equation}

\noindent which yields a standard single-agent actor-critic policy gradient $g=\mathbb{E}_{\mathbf{\pi}} [\nabla_{\theta} \log \mathbf{\pi}(\mathbf{u}|s) Q(s, \mathbf{u}) ]$. \cite{konda2000actor} establish that an actor-critic that follows this gradient converges to a local maximum of the
expected return $J^\pi$, subject to assumptions included in their paper.

In the naive critic framework, $V_{tot}(s)$ is evaluated by the central critic and does not depend on agent actions. Hence, by following the same proof in Equation \ref{eq2}, we can show that the expectation of naive critic baseline is also $0$, thus proves naive critic also converges to a locally optimal policy.



\begin{figure*}[ht!]

\begin{subfigure}{0.33\textwidth}
\includegraphics[width=\linewidth]{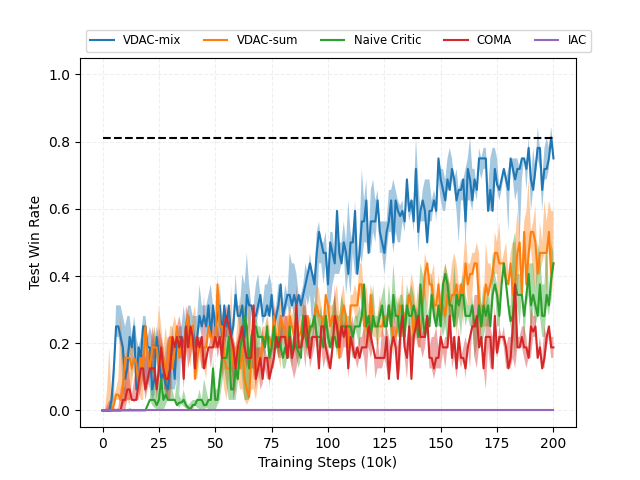}
\caption{1c3s5z}
\label{fig:1c3s5z}
\end{subfigure}
\begin{subfigure}{0.33\textwidth}
\includegraphics[width=\linewidth]{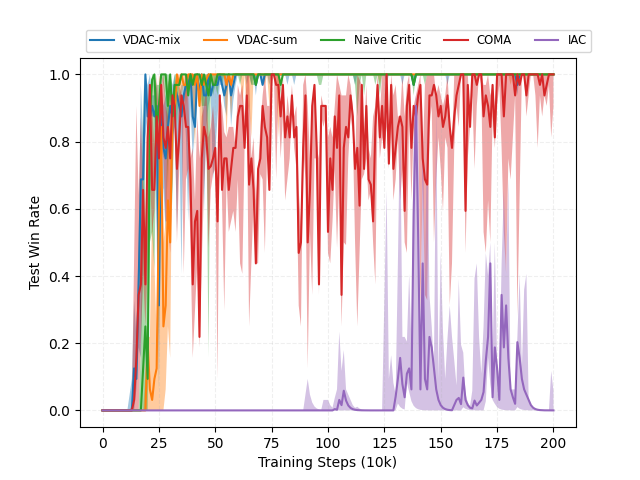} 
\caption{2s\_vs\_1sc}
\label{fig:2s_vs_1sc}
\end{subfigure}
\begin{subfigure}{0.33\textwidth}
\includegraphics[width=\linewidth]{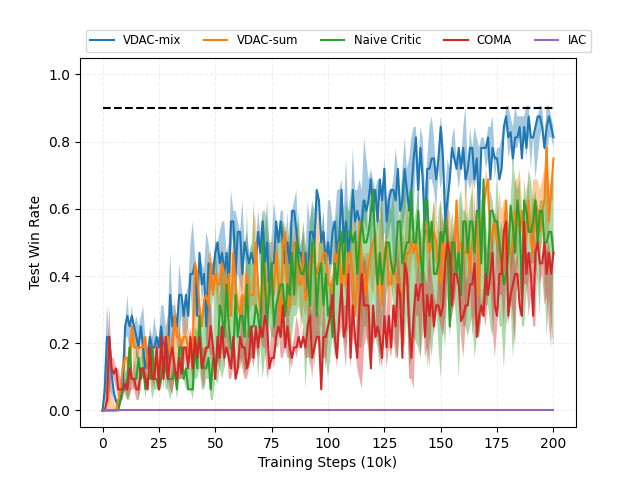}
\caption{2s3z}
\label{fig:2s3z}
\end{subfigure}
\begin{subfigure}{0.33\textwidth}
\includegraphics[width=\linewidth]{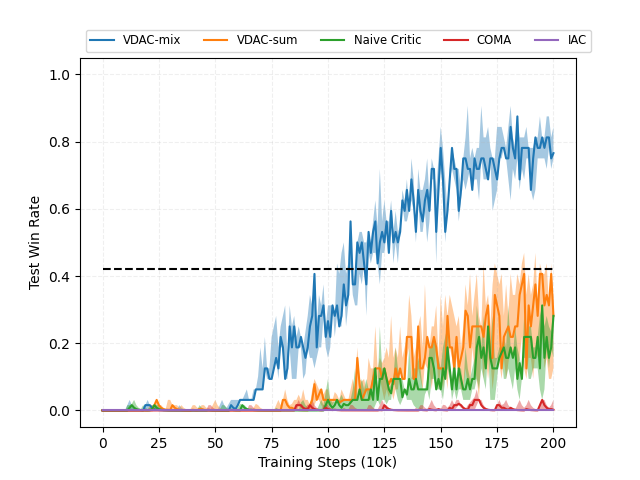}
\caption{3s5z}
\label{fig:3s5z}
\end{subfigure}
\begin{subfigure}{0.33\textwidth}
\includegraphics[width=\linewidth]{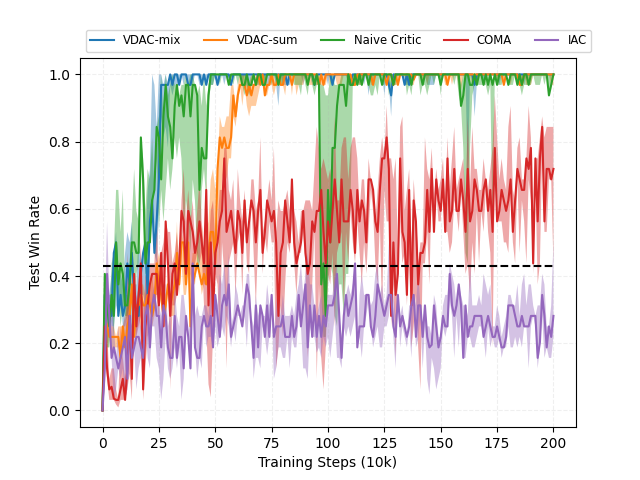} 
\caption{bane\_vs\_bane}
\label{fig:bane_vs_bane}
\end{subfigure}
\begin{subfigure}{0.33\textwidth}
\includegraphics[width=\linewidth]{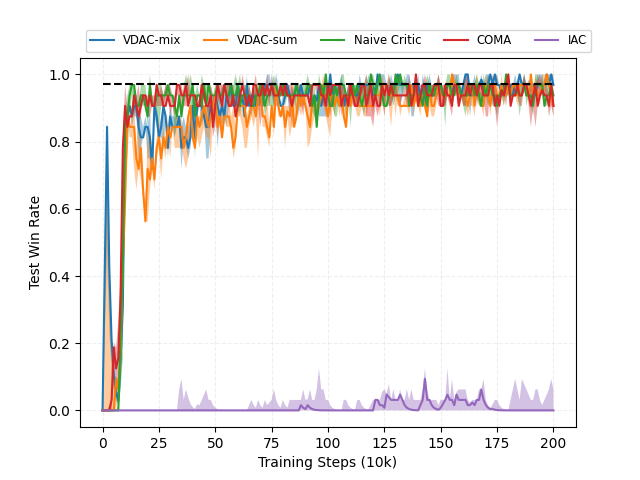}
\caption{8m}
\label{fig:8m}
\end{subfigure}

\caption{Overall results: Win rates on a range of SC mini-games. Black dash line represents heuristic AI's performance}
\label{fig:results}
\end{figure*}

\section{Experiments}\label{sec:exp}
In this section, we benchmark VDACs against the baseline algorithms listed in Table \ref{table: tj} on a standardized decentralised StarCraft II micromanagement environment, SMAC \cite{samvelyan2019starcraft}. SMAC consists of a set of StarCraft II micromanagement games that aim to evaluate how well independent agents are able to cooperate to solve complex tasks. In each scenario, algorithm-controlled ally units fight against enemy units controlled by the built-in game AI. An episode terminates when all units of either army have died or when the episode reached the pre-defined time limit. A game is counted as a win only if enemy units are eliminated. The goal is to maximize the win rate.

We consider the following maps in our experiments: 2s\_vs\_1sc, 2s3z, 3s5z, 1c3s5z, 8m, and bane\_vs\_bane. Note that all algorithms are trained under A2C framework where $8$ episodes are rolled out independently during the training. Refer to Appendix for training details and map configuration.

We perform the following ablations to answer the corresponding research questions: 
\subsubsection{Ablation 1} Is the TD advantage gradient sufficient to optimize multi-agent actor-critics?  The comparison between the naive critic and COMA will demonstrate the effectiveness of TD advantage policy gradients because the only significant difference between those two methods is that the naive critic follows a TD advantage policy gradient whereas COMA follows the COMA gradient (Equation \ref{eq: 4}).

\subsubsection{Ablation 2} Does applying state-value factorization improve the performance of actor-critic methods? VDAC-sum and IAC, both of which do not have access to extra state information, shares an identical structure. The only difference is that VDAC-sum applies a simple state-value factorization where the global state-value is a summation of local state values. The comparison between VDAC-sum and IAC will reveal the necessity of applying state-value factorization.

\subsubsection{Ablation 3} Compared with QMIX, does VDAC provide a reasonable trade-off between training efficiency and algorithm performance? We train VDAC and QMIX under A2C training paradigm, which is proposed to promote training efficiency, and compare their performance.

\subsubsection{Ablation 4} What are the factors that contribute to the performance of the proposed VDAC? We investigate the necessity of non-linear value-decomposition by removing the non-linear activation function in the mixing network. The resulting algorithm is called VDAC-mix (linear) and can be seen as VDAC-sum with access to extra state information. 



\subsection{Overall Results}

As suggested in \cite{samvelyan2019starcraft}, our main evaluation metric is the median win percentage of evaluation episodes as a function of environment steps observed over the $200$k training steps. Specifically, the performance of an algorithm is estimated by periodically running a fixed number of evaluation episodes (in our implementation, 32) during the course of training, with any exploratory behaviours disabled. The median performance as well as the $25$-$75\%$ percentiles are obtained by repeating each experiment using 5 independent training runs. Figure \ref{fig:results} demonstrates the comparison among actor-critics across 6 different maps.

In all scenarios, IAC fails to learn a policy that consistently defeats the enemy. In addition, its performance across training steps is highly unstable due to the non-stationarity of the environment and its lack of access to extra state information. 

Noticeably, VDAC-mix consistently achieves the best performance across all tasks. On easy games (i.e, 8m), all algorithms generally perform well. This is due to the fact that a simple strategy implemented by the heuristic AI to attack the nearest enemies is sufficient to win. In harder games such as 3s5z and 2s3z, only VDAC-mix can match or outperform the heuristic AI. It is worth noting that VDAC-sum, which cannot access extra state information, matches the naive critic's performance on most maps.



\begin{figure}
\centering
\includegraphics[scale=0.5]{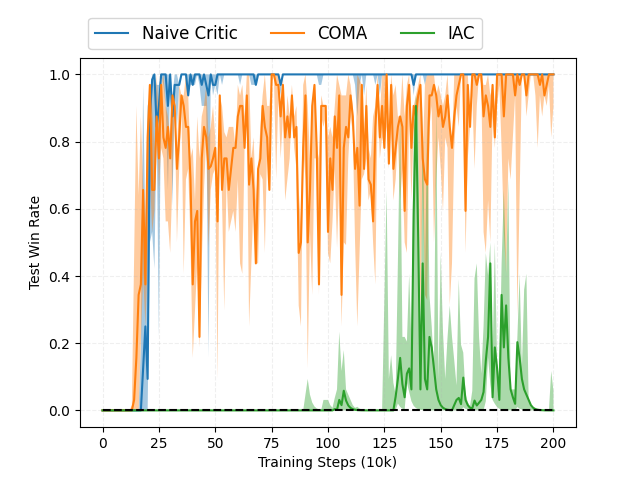} 
\caption{2s\_vs\_1sc (Ablation 1)}
\label{fig:nccoma_2s_vs_1sc}
\end{figure}

\begin{figure}
\centering
\includegraphics[scale=0.5]{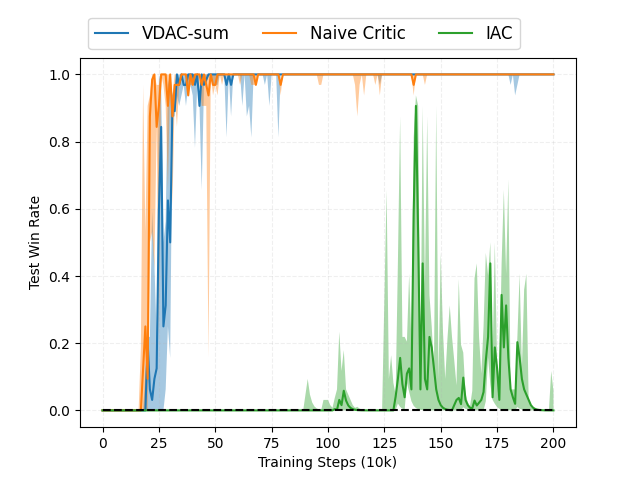} 
\caption{2s\_vs\_1sc (Ablation 2)}
\label{fig:vdniac_2s_vs_1sc}
\end{figure}

\begin{figure}
\centering
\includegraphics[scale=0.5]{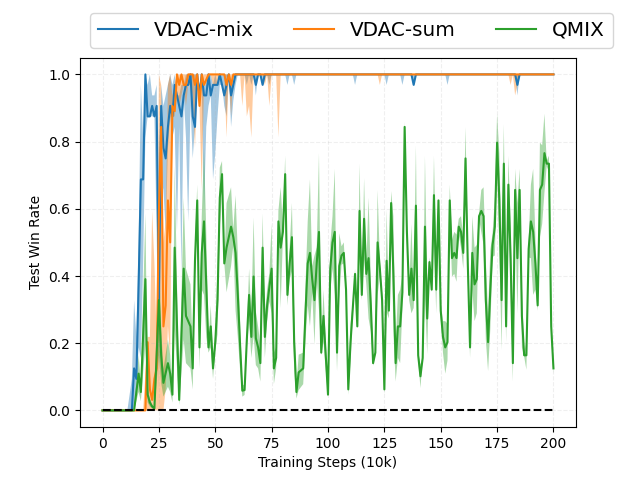} 
\caption{2s\_vs\_1sc (Ablation 3)}
\label{fig:VDACQMIX_2s_vs_1sc}
\end{figure}

\begin{figure}
\centering
\includegraphics[scale=0.5]{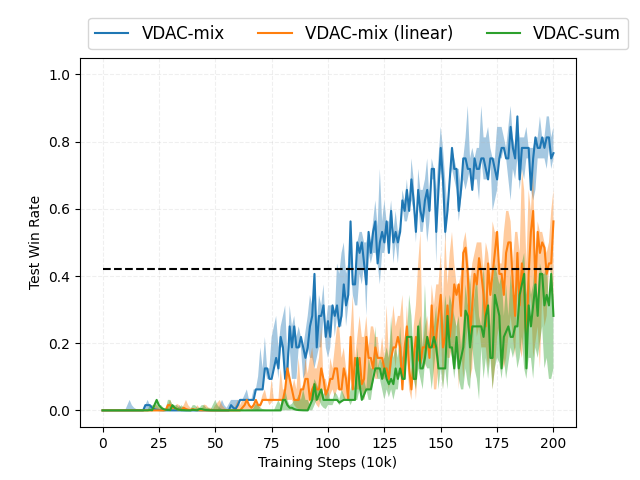} 
\caption{3s5z (Ablation 4)}
\label{fig:vmixlinear_3s5z}
\end{figure}

\subsubsection{Ablation 1}Consistent with \cite{lowe2017multi}, the comparison between the naive critic and IAC demonstrates the importance to incorporate extra state information, which is also revealed by the comparison between COMA and IAC (Refer to Figure \ref{fig:results} for comparisons between naive critic and COMA across different maps.). As shown in Figure \ref{fig:results}, naive critic outperforms COMA across all tasks. It reveals that it is also viable to use a TD advantage policy gradients in multi-agent settings. In addition, COMA's training is unstable, as can be seen in Figure \ref{fig:nccoma_2s_vs_1sc}, which might arise dues to its inability to predict accurate counterfactual action-value $Q^a(s, (\mathbf{u}^{-a}, u^a))$ for un-taken actions.


\subsubsection{Ablation 2} Despite the similarity in structure of VDAC-sum and IAC, VDAC-sum's median win rates at 2 million training step exceeds IAC's consistently across all maps (Refer to Figure \ref{fig:results} for comparisons between VDAC-sum and IAC across 6 different maps.). It reveals that, by using a simple relationship to enforce equation \ref{eq: mono}, we can drastically improve multi-agent actor-critic's performance. Furthermore, VDAC-sum matches naive critic on many tasks, as shown in Figure \ref{fig:vdniac_2s_vs_1sc}, demonstrating that actors that are trained without extra state information can achieve similar performance to naive critic by simply enforcing equation \ref{eq: mono}. In addition, it is noticeable that, compared with naive critic, VDAC-sum's performance is more stable across training. 



\subsubsection{Ablation 3}Figure \ref{fig:VDACQMIX_2s_vs_1sc} shows that, under the A2C training paradigm, VDAC-mix outperforms QMIX in map 2s\_vs\_1sc. It is also noticeable that QMIX's performance is unstable across the training steps in map 2s\_vs\_1sc. In easier games, QMIX's performance can be comparable to VDAC-mix. In harder games such as 3s5z, VDAC-mix's median test win rates at 2 million training step outnumber QMIX's by 71\%. Refer to Appendix for complete comparisons between VDACs and QMIX.


\subsubsection{Ablation 4}Finally, we introduced VDAC-mix (linear), which can be seen as a more general VDAC-sum that has access to extra state information. Consistent with our previous conclusion, the comparison between VDAC-mix (linear) and VDAC-sum shows that it is important to incorporate extra state information. In addition, the comparison between VDAC-mix and VDAC-mix (linear) shows the necessity of assuming the non-linear relationship between the global state value $V_{tot}$ and local state values $V^a, \forall a \in \{1,\dots,n\}$. Refer to Appendix for comparisons between VDACs across all maps.





\section{Conclusion}\label{sec:conclusions}
In this paper, we propose a new credit-assignment actor-critic framework that enforces the monotonic relationship between the global state-value and the shaped local state-value. Theoretically, we establish the convergence of the proposed actor-critic method to a local optimal. Empirically, benchmark tests on StarCraft micromanagement games demonstrate that our proposed actor-critic bridges the performance gap between multi-agent actor-critics and Q-learning, and our method provides a balanced trade-off between training efficiency and performance. Furthermore, we identify a set of key factors that contribute to the performance of our proposed algorithms via a set of ablation experiments. In future work, we aim to implement our framework in real-world applications such as highway on-ramp merging of semi or full self-driving vehicles.

\clearpage

\section{Ethical Impact of Work}

Large-scale multi-agent control problems are at the heart of a number of challenging problems facing society. For example in traffic management, there are over 300,000 accidents per year that occur during highway merging.  The number of accidents could be significantly reduced if effective autonomous driving was broadly available in personal vehicles. MARL algorithms, like ones proposed in this paper, offer a possible solution to the autonomous driving task.  Other areas of significant societal impact include healthcare, smart manufacturing, smart grids, and other transportation infrastructure.

%
%
%
\bibliography{mypaper.bib}
\appendix
\onecolumn
\section{Appendix}
\subsection{SMAC}
In this paper, we use all the default settings in \cite{samvelyan2019starcraft}. That includes: the game difficulty is set to level $7$, \textit{very difficult}, the shoot range, observe range, etc, are consistent with the default settings. The action space of agents consists of the following set of discrete actions: move[direction], attack[enemy id], stop, and no operation. Agents can only move in four directions: north, south, east, or west. A unit is allowed to perform the attack[enemy id] action only if the enemy is within its shooting range.

Each unit has a sight range that limits its ability to receive any information out of range. The sight range, which is bigger than shooting range, makes the environment partially observable from the standpoint of each agent. Agents can only observe other agents if they are both alive and located within the sight range. The global state, which is only available to agents during centralised training, encapsulates information about all units on the map. 

The observation vector also follows the default implementation in \cite{samvelyan2019starcraft}: It contains the following attributes for both allied and enemy units within the sight range: distance, relative x, relative y, health, shield, and unit type. In addition, the observation vector includes the last actions of allied units that are in the field of view. Lastly, the terrain features, in particular the values of eight points at a fixed radius indicating height and walkability, surrounding agents within the observe range are also included. The state vector includes the coordinates of all agents relative to the center of the map, together with units' observation feature vectors. Additionally, the energy of Medivacs and cooldown of the rest of the allied units are stored in the state vector. Finally, the last actions of all agents are attached to the state vector.

\begin{table*}[ht!] 
\centering
\caption{Map Descriptions.}
\begin{tabular}{c l l c l c}
\hline
Map Name &  & &Ally Units & & Enemy Units\\
\hline
\hline
2s\_vs\_1sc & & &2 Stalkers & &1 Spine Crawler \\
8m & & &8 Marines & &8 Marines \\
2s3z & & &2 Stalkers \& 3 Zealots & &2 Stalkers \& 3 Zealots \\
3s5z & & &3 Stalkers \& 5 Zealots & &3 Stalkers \& 5 Zealots\\
1c3s5z & & &1 Colossus, 3 Stalkers \& 5 Zealots & &1 Colossus, 3 Stalkers \& 5 Zealots\\
bane\_vs\_bane & & &20 Zerglings \& 4 Banelings & &20 Zerglings \& 4 Banelings\\
\hline

\label{table: sc_desc}
\end{tabular}
\end{table*}

\subsection{Training Details and Hyperparameters}
Experiments are obtained by using Nvidia RTX 2080 Ti graphics cards, with each independent run taking 1 to 3 hours depending on the scenario. Each independent run corresponds to a unique random seed that is randomly generalized at the beginning.

The agent networks of all algorithms resemble a DRQN \cite{hausknecht2015deep} with a recurrent layer comprised of a GRU \cite{chung2014empirical} with a 64-dimensional hidden state, with a fully-connected layer before and after. The exception is that IAC, VDAC-sum, and VDAC-mix agent networks contain an additional layer to output local state values and the policy network outputs a stochastic policy rather than action-values.

Algorithms are trained with RMSprop with learning rate $5\times 10^{-4}$. During training, $8$ games are initiated independently, from which episodes are sampled. Q-learning replay buffer stores the latest 5000 episodes for each independent game (In total, replay buffer has a size of $8\times5000=40000$). We set $\gamma=0.99$ and $\lambda=0.8$ (if needed). Target networks (if exists) are updated every $200$ training steps.

The architecture of the COMA critic is a feedforward fully-connected neural network with the first $2$ layers, each of which has $128$ units, followed by a final layer of $|U|$ units. Naive central critic shares the same architecture with COMA critic with an exception that its final layer contains $1$ units.

The mixing network in QMIX and VDAC-mix shares an identical structure. It consists of a single hidden layer of 32 units, whose parameters are outputted by hypernetworks. An ELU activation function follows the hidden layer in the mixing network. The hypernetworks consist of a feedforward network with a single hidden layer of 64 units with a ReLU activation function.

For naive central critic, IAC, and VDACs, $Q(s_t,\mathbf{u}_t)$ is given by:
\begin{equation}
    Q(s_t,\mathbf{u}_t) = \sum_{i=0}^{k-1} \gamma^i r_{t+i} + \gamma^k V(s_{t+k}),
\end{equation} where $k$ can vary from state to state and is upper-bounded by $T$.

\begin{algorithm*}[ht!] \label{alg: cc}
	\caption{Naive Central Critic} 
	\begin{algorithmic}[1]
	\State Initialize critic $\theta^c$, target critic $\hat{\theta}^c$, and actor $\theta$
		\For {each training episode $e$}
		    \State Empty buffer
			\For {$e_c=1$ to $\frac{\text{BatchSize}}{n}$}
				\State $t=0, h_o^a$ for each agent $a$
				\While {game not terminated \textbf{and} $t<T$} 
				    \State $t=t+1$
				    \For {each agent $a$}
				        \State $h^a_t, \pi^a_t=\text{Actor}(o_t^a,h_{t-1}^a,u^a_{t-1},a;\theta)$
				        \State Sample action $u^a_t$ from $\pi^a_t$
				    \EndFor
				    \State Get reward $r_t$ and next state $s_{t+1}$
			    \EndWhile
				\State add experience to buffer
			\EndFor
			\State Collate episodes in buffer into single batch
			\For {$t=1$ to $T$}
			    \State Batch unroll RNN using states, actions and reward
			    \State Calculate $y_t$ and $A_t$ using $\hat{\theta}^c$
			\EndFor
			
			\For {$t=T$ down to $1$}
			    \State Calculate gradient wrt $\theta^{c}: \Delta\theta^{c} \leftarrow \nabla_{\theta^{c}}\big(y_t-V(s_t,\mathbf{u}_{t-1};\theta^{c})\big)^2$
			    \State Update critic $\theta^c \leftarrow \theta^c - \alpha \Delta\theta^{c}$
			    \State Every C steps update target critic $\hat{\theta^c} \leftarrow \theta^c$
    		\EndFor
    		\For {$t=1$ down to $T$}
			    \State Accumulate gradient wrt $\theta: \Delta\theta \leftarrow \Delta\theta + \nabla_{\theta}\log \pi(u_t^a|o_t^a)A_t$
    		\EndFor
    		
    		\State Update actor weights $\theta=\theta+\alpha \Delta\theta$
		\EndFor
	\end{algorithmic} 
\end{algorithm*}

\begin{algorithm*}[ht!] \label{alg: ppo_sum}
	\caption{Value Decomposition Actor-Critic (VDAC-sum)} 
	\begin{algorithmic}[1]
	\State Initialize actor network $\theta$
		\For {each training episode $e$}
		    \State Empty buffer
			\For {$e_c=1$ to $\frac{\text{BatchSize}}{n}$}
				\State $t=0, h_o^a$ for each agent $a$
				\While {game not terminated \textbf{and} $t<T$} 
				    \State $t=t+1$
				    \For {each agent $a$}
				        \State $h^a_t, \pi^a_t, V^a_t=\text{Actor}(o_t^a,h_{t-1}^a,u^a_{t-1},a;\theta)$
				        \State Sample action $u^a_t$ from $\pi^a_t$
				    \EndFor
				    \State Get reward $r_t$ and next state $s_{t+1}$
			    \EndWhile
				\State add experience to buffer
			\EndFor
			\State Collate episodes in buffer into single batch
			\For {$t=1$ to $T$}
			    \State Batch unroll RNN using states, actions and reward
			    \State Calculate $y_t$ and $A_t$ using ${\theta}$
			    \State Accumulate gradient wrt $\theta: \Delta\theta_v \leftarrow \Delta\theta_v+\nabla_{\theta}\big(y_t-\sum_aV^a_{t}\big)^2$
			\EndFor
			
			\For {$t=1$ to $T$}
			    \State Accumulate gradient wrt $\theta: \Delta\theta_{\pi} \leftarrow \Delta\theta_{\pi} + \nabla_{\theta}\log \pi(u_t^a|o_t^a)A_t$
    		\EndFor
    		\State Update actor weights $\theta=\theta+\alpha_{\pi} \Delta\theta_{\pi}-\alpha_{v} \Delta\theta_{v}$
		\EndFor
	\end{algorithmic} 
\end{algorithm*}
\begin{algorithm*}[ht!] \label{alg: ppo}
	\caption{Value Decomposition Actor-Critic (VDAC-mix)} 
	\begin{algorithmic}[1]
	\State Initialize hypernetwork $\theta^c$, and actor network $\theta$
		\For {each training episode $e$}
		    \State Empty buffer
			\For {$e_c=1$ to $\frac{\text{BatchSize}}{n}$}
				\State $t=0, h_o^a$ for each agent $a$
				\While {game not terminated \textbf{and} $t<T$} 
				    \State $t=t+1$
				    \For {each agent $a$}
				        \State $h^a_t, \pi^a_t, V^a_t=\text{Actor}(o_t^a,h_{t-1}^a,u^a_{t-1},a;\theta)$
				        \State Sample action $u^a_t$ from $\pi^a_t$
				    \EndFor
				    \State Get reward $r_t$ and next state $s_{t+1}$
			    \EndWhile
				\State add experience to buffer
			\EndFor
			\State Collate episodes in buffer into single batch
			\For {$t=1$ to $T$}
			    \State Batch unroll RNN using states, actions and reward
			    \State Calculate $y_t$ and $A_t$ using ${\theta}^c$
			    \State Accumulate gradient wrt $\theta^c:\Delta \theta^c \leftarrow \Delta\theta^c+\nabla_{\theta^c}\big(y_t-V_{tot}(V^1_{t},\dots,V^{n}_t)\big)^2$
			    \State Accumulate gradient wrt $\theta: \Delta\theta_v \leftarrow \Delta\theta_v+\nabla_{\theta}\big(y_t-V_{tot}(V^1_{t},\dots,V^{n}_t)\big)^2$
			\EndFor
			
			\For {$t=1$ to $T$}
			    \State Accumulate gradient wrt $\theta: \Delta\theta_{\pi} \leftarrow \Delta\theta_{\pi} + \nabla_{\theta}\log \pi(u_t^a|o_t^a)A_t$
    		\EndFor
    		\State Update actor weights $\theta=\theta+\alpha_{\pi} \Delta\theta_\pi-\alpha_{v} \Delta\theta_v$
    		\State Update hypernet weights $\theta^{c}=\theta^{c}-\alpha \Delta\theta^c$
		\EndFor
	\end{algorithmic} 
\end{algorithm*}

\subsection{StarCraft II Results}
\begin{figure*}[ht!]

\begin{subfigure}{0.33\textwidth}
\includegraphics[width=\linewidth]{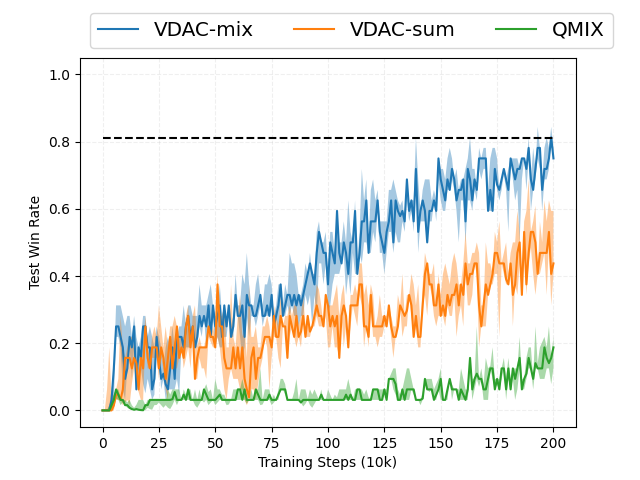}
\caption{1c3s5z}
\label{fig:VDAC(QMIX)_1c3s5z(a)}
\end{subfigure}
\begin{subfigure}{0.33\textwidth}
\includegraphics[width=\linewidth]{VDACQMIX_2s_vs_1sc.png} 
\caption{2s\_vs\_1sc}
\label{fig:VDAC(QMIX)_2s_vs_1sc(a)}
\end{subfigure}
\begin{subfigure}{0.33\textwidth}
\includegraphics[width=\linewidth]{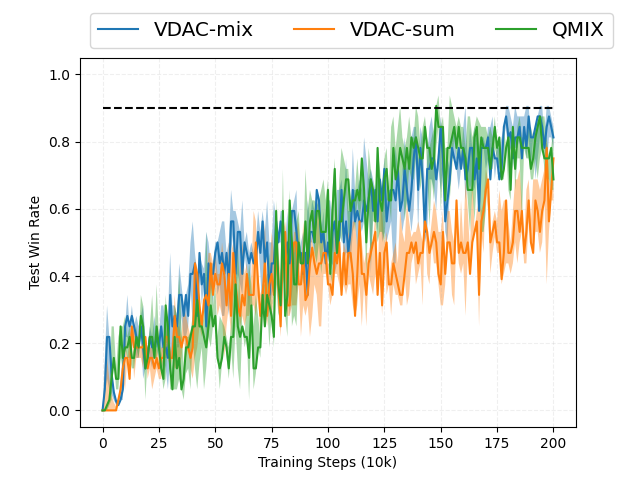}
\caption{2s3z}
\label{fig:VDAC(QMIX)_2s3z(a)}
\end{subfigure}
\begin{subfigure}{0.33\textwidth}
\includegraphics[width=\linewidth]{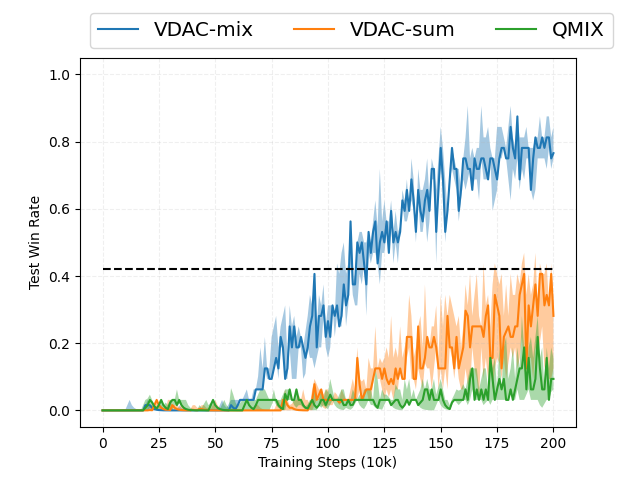}
\caption{3s5z}
\label{fig:VDAC(QMIX)_3s5z(a)}
\end{subfigure}
\begin{subfigure}{0.33\textwidth}
\includegraphics[width=\linewidth]{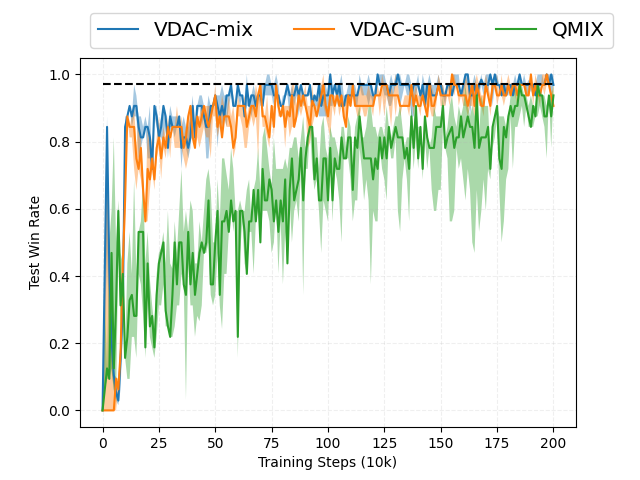}
\caption{8m}
\label{fig:VDAC(QMIX)_8m}
\end{subfigure}

\caption{Overall results: VDACs vs QMIX under A2C}
\label{fig:VDAC(QMIX)}
\end{figure*}

\begin{figure*}[ht!]

\begin{subfigure}{0.33\textwidth}
\includegraphics[width=\linewidth]{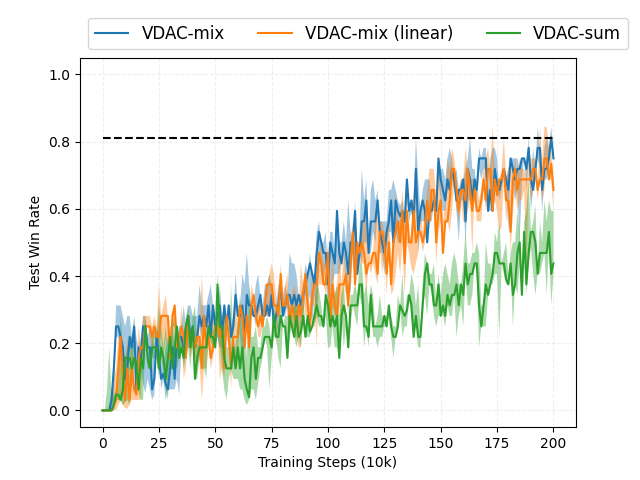}
\caption{1c3s5z}
\label{fig:vmix(linear)_1c3s5z(a)}
\end{subfigure}
\begin{subfigure}{0.33\textwidth}
\includegraphics[width=\linewidth]{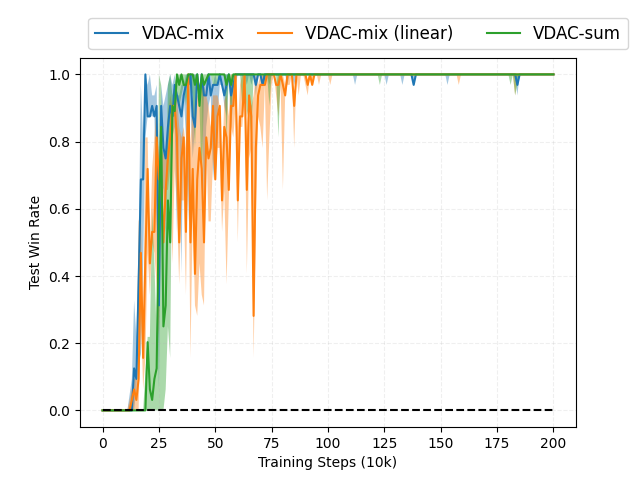} 
\caption{2s\_vs\_1sc}
\label{fig:vmix(linear)_2s_vs_1sc(a)}
\end{subfigure}
\begin{subfigure}{0.33\textwidth}
\includegraphics[width=\linewidth]{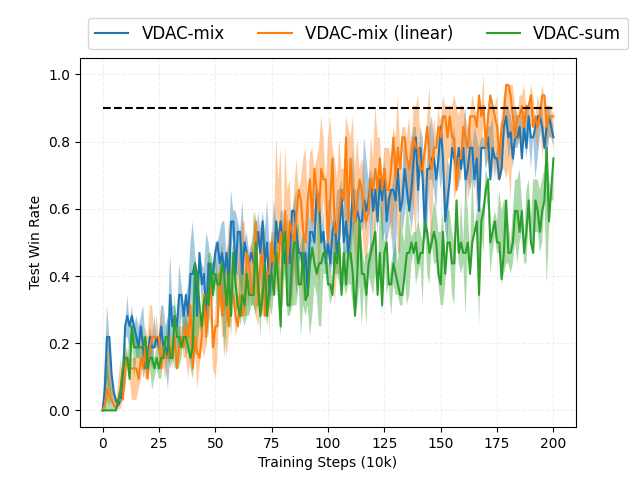}
\caption{2s3z}
\label{fig:vmix(linear)_2s3z(a)}
\end{subfigure}
\begin{subfigure}{0.33\textwidth}
\includegraphics[width=\linewidth]{vmixlinear_3s5z.png}
\caption{3s5z}
\label{fig:vmix(linear)_3s5z(a)}
\end{subfigure}
\begin{subfigure}{0.33\textwidth}
\includegraphics[width=\linewidth]{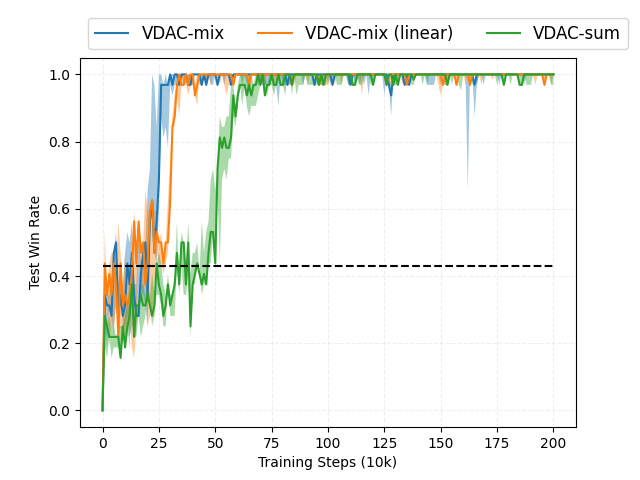} 
\caption{bane\_vs\_bane}
\label{fig:vim(linear)_bane_vs_bane(a)}
\end{subfigure}
\begin{subfigure}{0.33\textwidth}
\includegraphics[width=\linewidth]{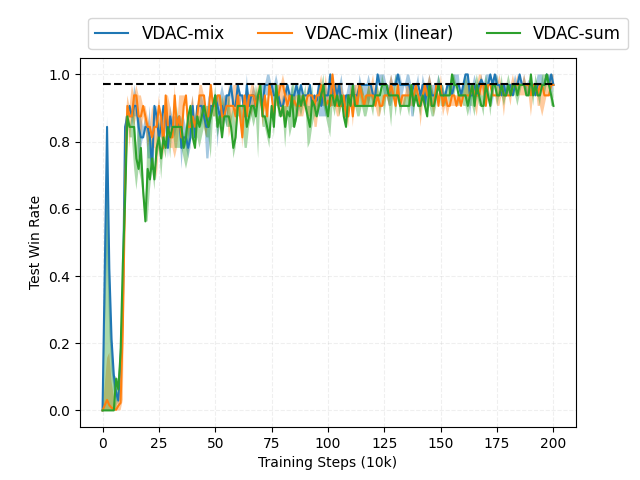}
\caption{8m}
\label{fig:vmix(linear)_8m(a)}
\end{subfigure}

\caption{Overall results: VDAC-mix vs VDAC-mix(linear) vs VDAC-sum}
\label{fig:VDAC(linear)}
\end{figure*}
\end{document}